\documentclass[times, twoside]{zHenriquesLab-StyleBioRxiv}
\usepackage{times}
\usepackage{epsfig}
\usepackage{graphicx}
\usepackage{amsmath}
\usepackage{amssymb}
\usepackage{csquotes}
\usepackage[percent]{overpic}
\usepackage{multirow}
\usepackage{bm}
\usepackage[export]{adjustbox}
\usepackage{xspace}

\makeatletter
\DeclareRobustCommand\onedot{\futurelet\@let@token\@onedot}
\def\@onedot{\ifx\@let@token.\else.\null\fi\xspace}
\def\eg{\emph{e.g}\onedot} 
\def\ie{\emph{i.e}\onedot}

\def\wrt{w.r.t\onedot} 
\def\etal{\emph{et~al}\onedot}
\makeatother

\newcommand{\compactsubsub}[2]{\vspace{1.2mm}\noindent\textit{\textbf{#1:}} #2}

\leadauthor{Buchholz} 
\newcommand\figTeaser{
\begin{figure}[t]
    \centering
    \begin{overpic}[width=\linewidth, trim=170 50 240 4, clip]
    {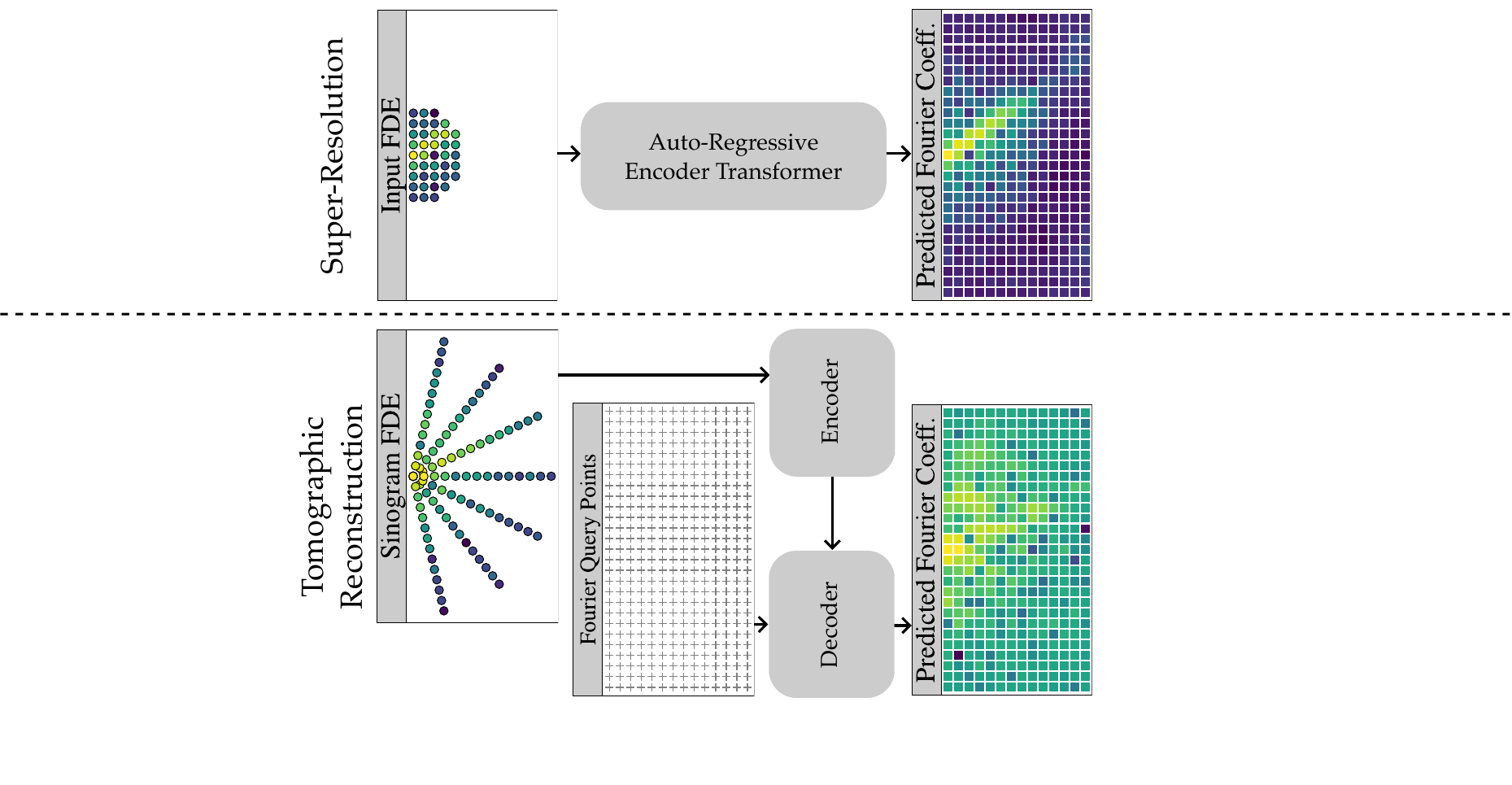}
    \end{overpic}
    \caption{We present Fourier Image Transformers (FITs), realizations of Fast-Transformers~\cite{katharopoulos_et_al_2020}, that operate on images via a novel sequential image representation we call Fourier Domain Encoding (FDE).
    We demonstrate the utility of FITs on two tasks.
    First, an image super-resolution task \textit{(top row)}, where the low frequencies of an image are used to predict missing high frequencies, corresponding to a higher resolution image.
    The second task is tomographic reconstruction \textit{(bottom row)}, where we provide a sparse sampled Fourier space as input to an encoder-decoder FIT and predict missing Fourier coefficients.
    Note that the real images can be obtained from both predictions by taking the inverse Fourier transform.}
    \label{fig:teaser}
\end{figure}
}

\newcommand\figSResOverview{
\begin{figure}[t]
    \centering
    \begin{overpic}[trim=1 130 0 1, clip, width=\linewidth]
    {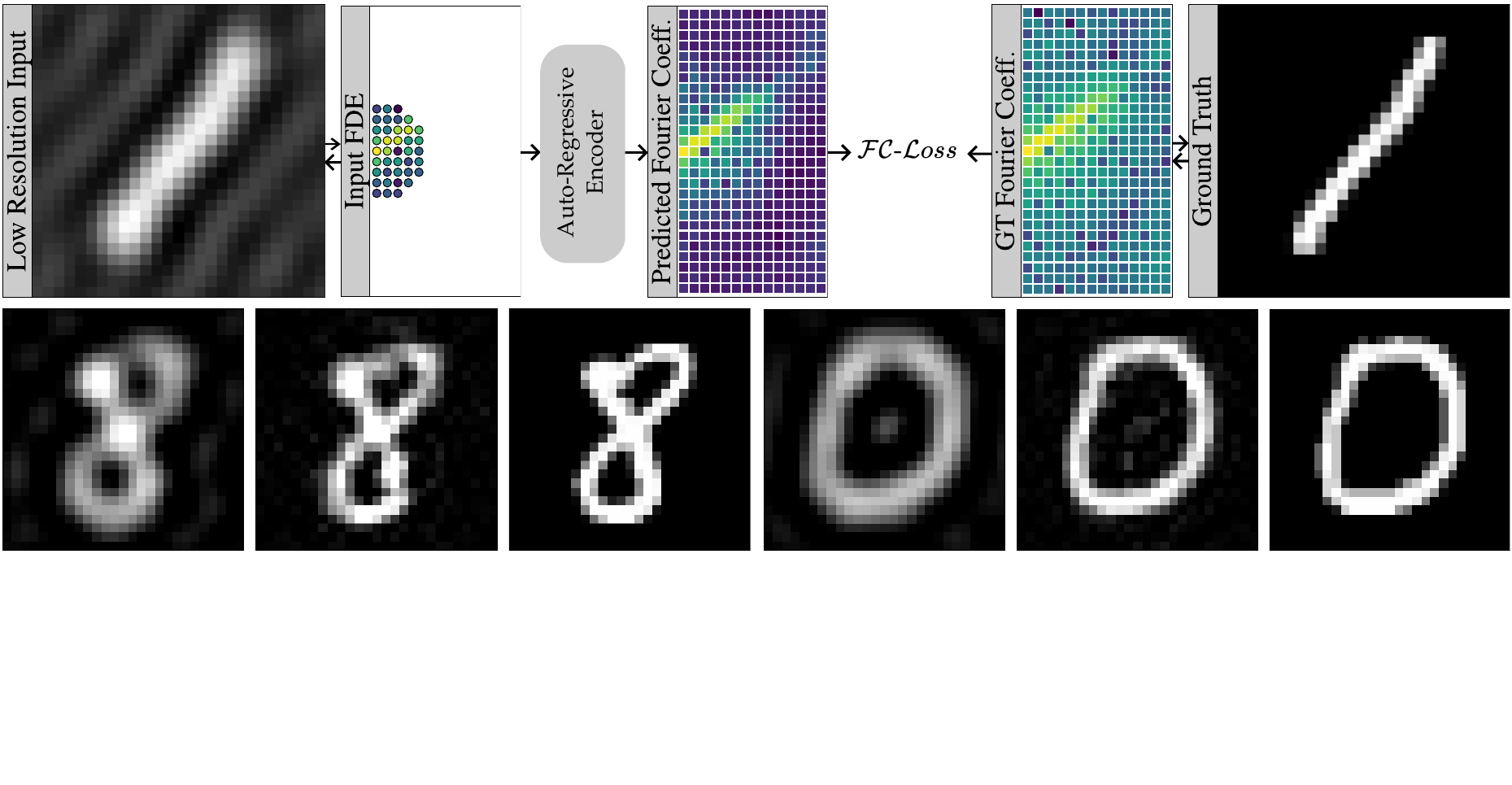}
    \put(0.5,2){\color{white}\scriptsize{(a)}}
    \put(17.2,2){\color{white}\scriptsize{(b)}}
    \put(34.1,2){\color{white}\scriptsize{(c)}}
    \put(51.0,2){\color{white}\scriptsize{(d)}}
    \put(67.6,2){\color{white}\scriptsize{(e)}}
    \put(84.5,2){\color{white}\scriptsize{(f)}}
    \end{overpic}
    \caption{\textbf{FIT for super-resolution.}
    Low-resolution input images are first transformed into Fourier space and then unrolled into an FDE sequence, as described in Section~\ref{sec:fde}.
    This FDE sequence can now be fed to a FIT, that, conditioned on this input, extends the FDE sequence to represent a higher resolution image.
    This setup is trained using an $\mathcal{FC}\text{-}\mathcal{L}oss$ that enforces consistency between predicted and ground truth Fourier coefficients.
    During inference, the FIT is conditioned on the first $39$ entries of the FDE, corresponding to \textit{(a,d)}~$3\times$ Fourier binned input images. 
    Panels \textit{(b,e)} show the inverse Fourier transform of the predicted output, and panels
    \textit{(c,f)}~depict the corresponding ground truth.
    }    
    \label{fig:sres_overview}
\end{figure}
}

\newcommand\figTRecOverview{
\begin{figure}[th]
    \centering
    \begin{overpic}[width=\linewidth, trim=0 44 0 0, clip]{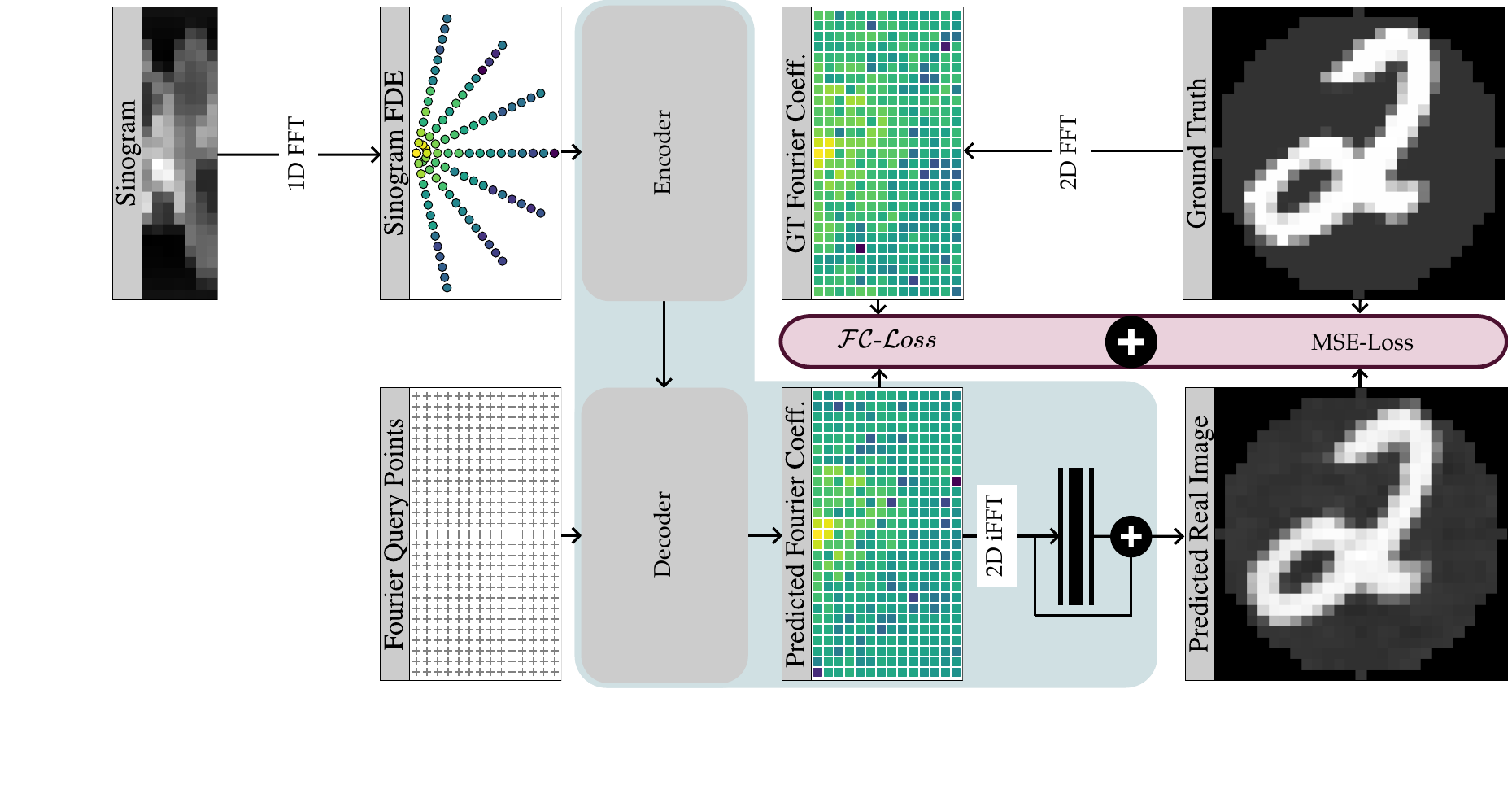}
    \put(0.5,44){\color{black}\scriptsize{(a)}}
    \end{overpic}
    \begin{overpic}[width=\linewidth, trim=0 54 0 0, clip]{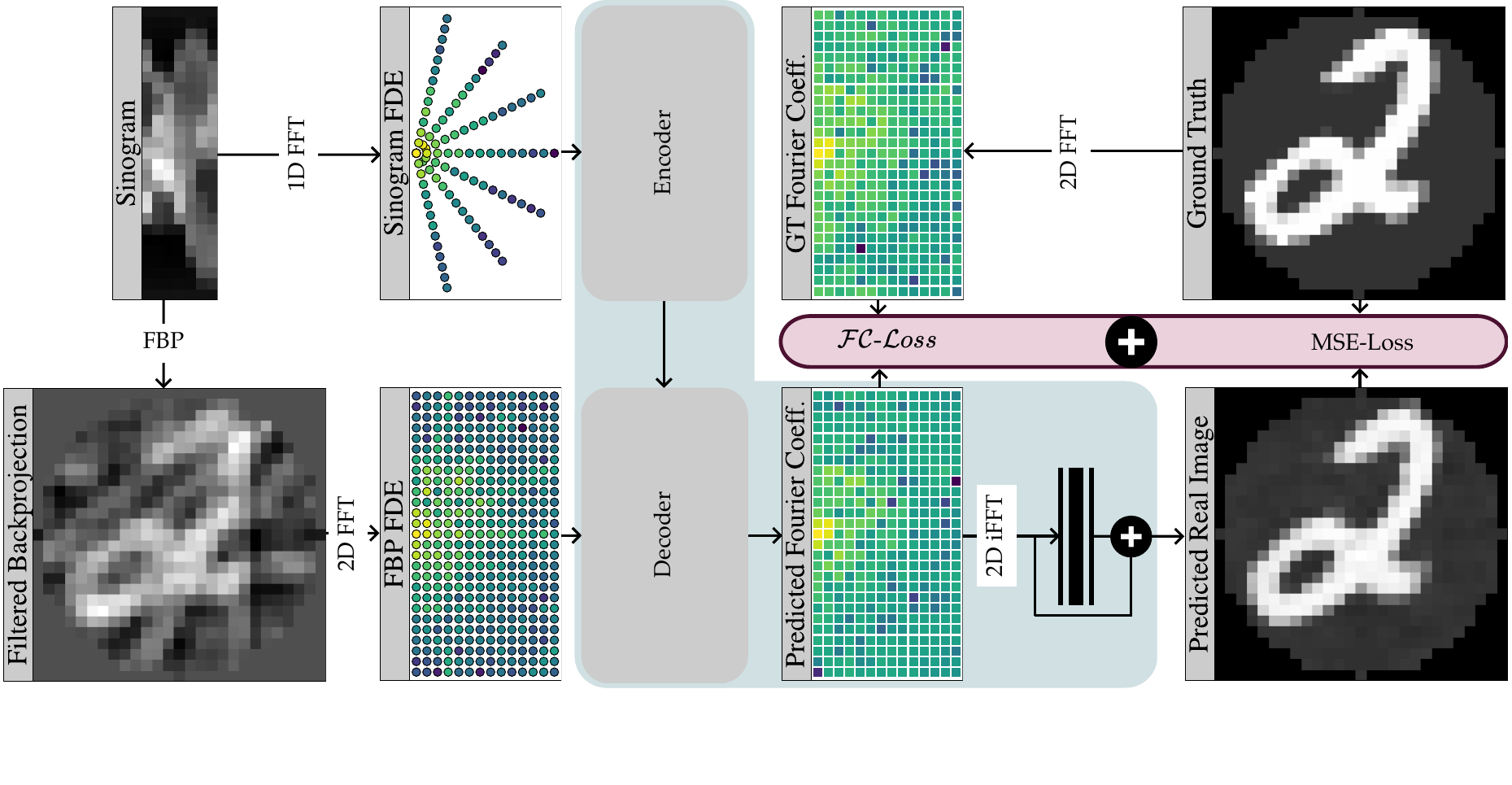}
    \put(0.5,43){\color{black}\scriptsize{(b)}}
    \end{overpic}

    \caption{\textbf{FIT for computed tomography.}
    In 2D computed tomography, 1D projections of an imaged sample (\ie the columns of a sinogram) are back-transformed into a 2D image.
    We propose two encoder-decoder based Fourier Image Transformer setups for tomographic reconstruction. 
    For ``FIT: TRec'', subfigure~$\bm{(a)}$, the sinogram is converted into our proposed FDE which serves as input to the encoder. The decoder predicts a dense Fourier spectrum from the latent encoding and provided Fourier query points. 
    In order to reduce high frequency fluctuations in this result, we introduce a shallow conv-block after the iFFT  (shown in black).
    The setup is trained with the $\mathcal{FC}\text{-}\mathcal{L}oss$, see Section~\ref{sec:fc_loss}, and a conventional MSE-loss between prediction and ground truth.
    For ``FIT: TRec + FBP'', sub-figure~$\bm{(b)}$, we enrich the Fourier query points with encoded information from the filtered backprojection (FBP)~\cite{kak2002principles,ramesh1989algorithm}.
    }
    \label{fig:trec_overview}
\end{figure}
}

\newcommand\figSResQuant{
\begin{figure*}[t]
    \centering
    \begin{overpic}[width=\linewidth]
    {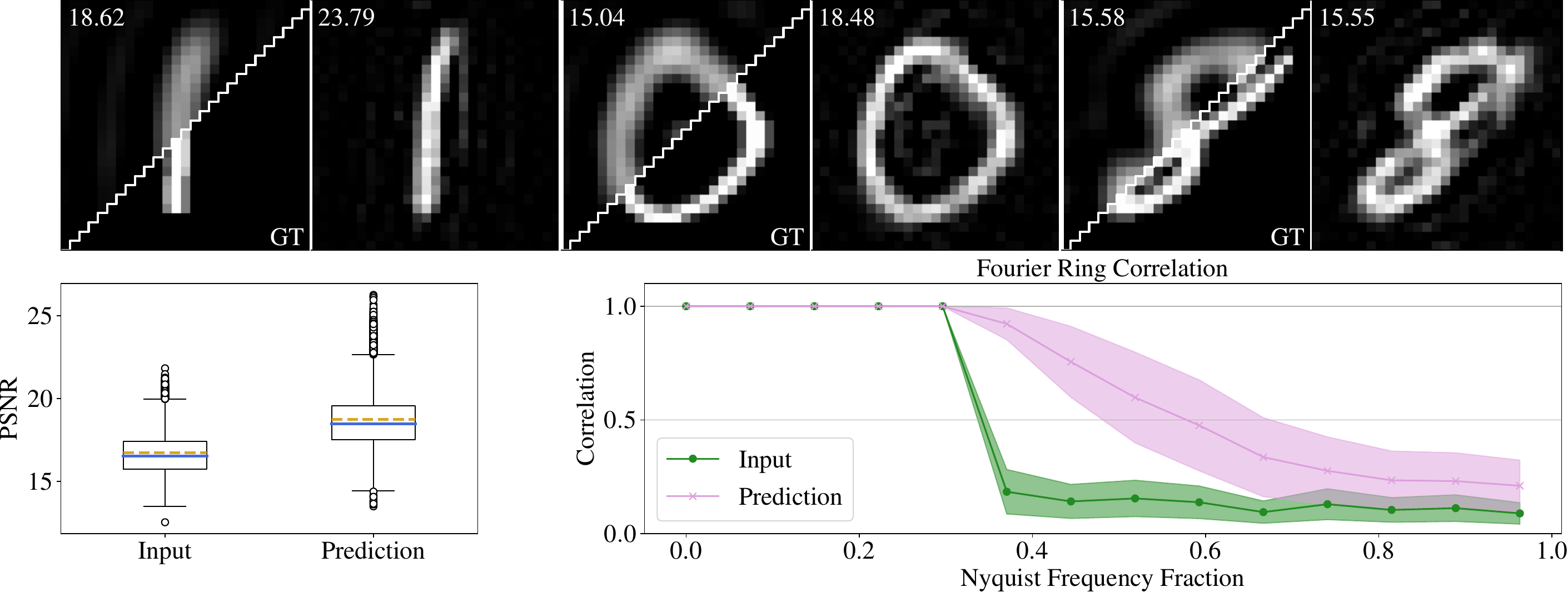}
    \put(4,22.4){\colorbox{black}{\makebox(1,1){\color{white}(a)}}}
    \put(20,22.4){\colorbox{black}{\makebox(1,1){\color{white}(b)}}}
    \put(36,22.4){\colorbox{black}{\makebox(1,1){\color{white}(c)}}}
    \put(51.9,22.4){\colorbox{black}{\makebox(1,1){\color{white}(d)}}}
    \put(67.85,22.4){\colorbox{black}{\makebox(1,1){\color{white}(e)}}}
    \put(83.8,22.4){\colorbox{black}{\makebox(1,1){\color{white}(f)}}}
    \end{overpic}
    \caption{\textbf{Super-resolution results on MNIST.}
    The top left triangles of \textit{(a,c,e)} show $3\times$ binned (low-res) MNIST inputs. 
    Conditioned by these inputs, our trained FIT auto-regressively generates results as shown in \textit{(b,d,f)}. 
    Inputs and predictions are labeled with the peak signal-to-noise ratio (PSNR) values computed \wrt ground truth images, shown in the lower-right half of \textit{(a,c,e)}. 
    The examples in \textit{(a-f)} correspond to the 98th, 50th and 2nd percentile in terms of obtained PSNR over all MNIST prediction results.
    Box-plots show the distribution of PSNR values of Fourier binned inputs and predicted outputs, respectively (mean in dashed gold and median in solid blue). 
    The Fourier ring correlation plot shows how predicted Fourier coefficients are improving \wrt true ground truth coefficients.
    Shaded areas correspond to $+/-1$ standard deviation. 
    The correlation for the first $5$ Fourier shells is $1$ because these shells have been used as inputs to the FIT.
    }
    \label{fig:sres_quant}
\end{figure*}
}

\newcommand\figSResCond{
\begin{figure}[t]
    \centering
    \begin{overpic}[width=\linewidth]
    {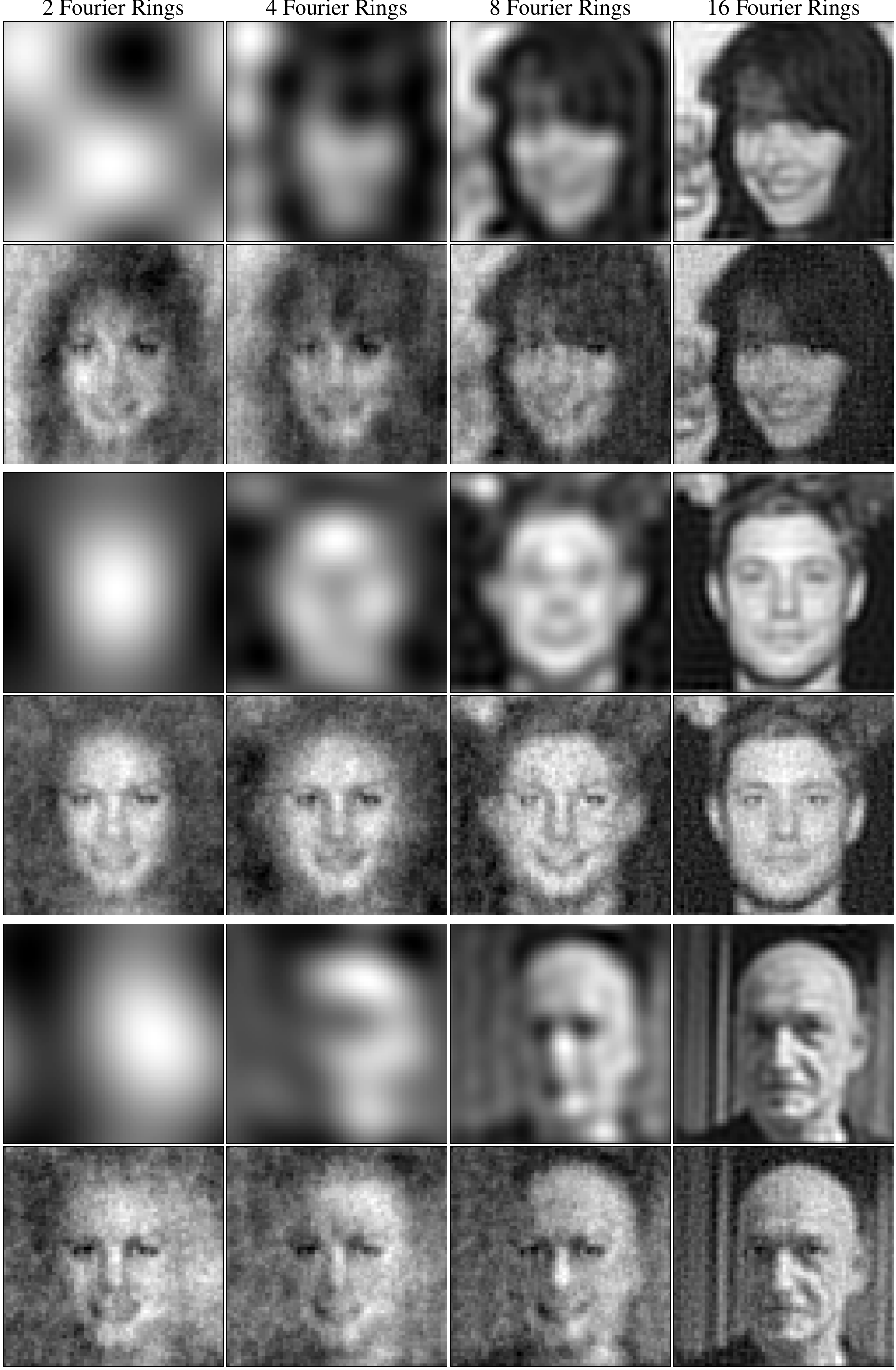}
    \end{overpic}
    \caption{\textbf{Super-resolution results on CelebA.}
    On $3$ input images (rows $1$+$2$, $3$+$4$, and $5$+$6$) we show predictions of a trained super-resolution FIT conditioned on $2$, $4$, $8$ and $16$ Fourier rings (columns).
    Upper rows show the iFFT of the input FDEs used to condition the FIT, lower rows depict the iFFT of the predicted results.}
    \label{fig:sres_cond}
\end{figure}
}

\newcommand\figTRecQuali{
\begin{figure}[t]
    \begin{overpic}[width=.97\linewidth, trim=0 1083 0 0, clip,right]
    {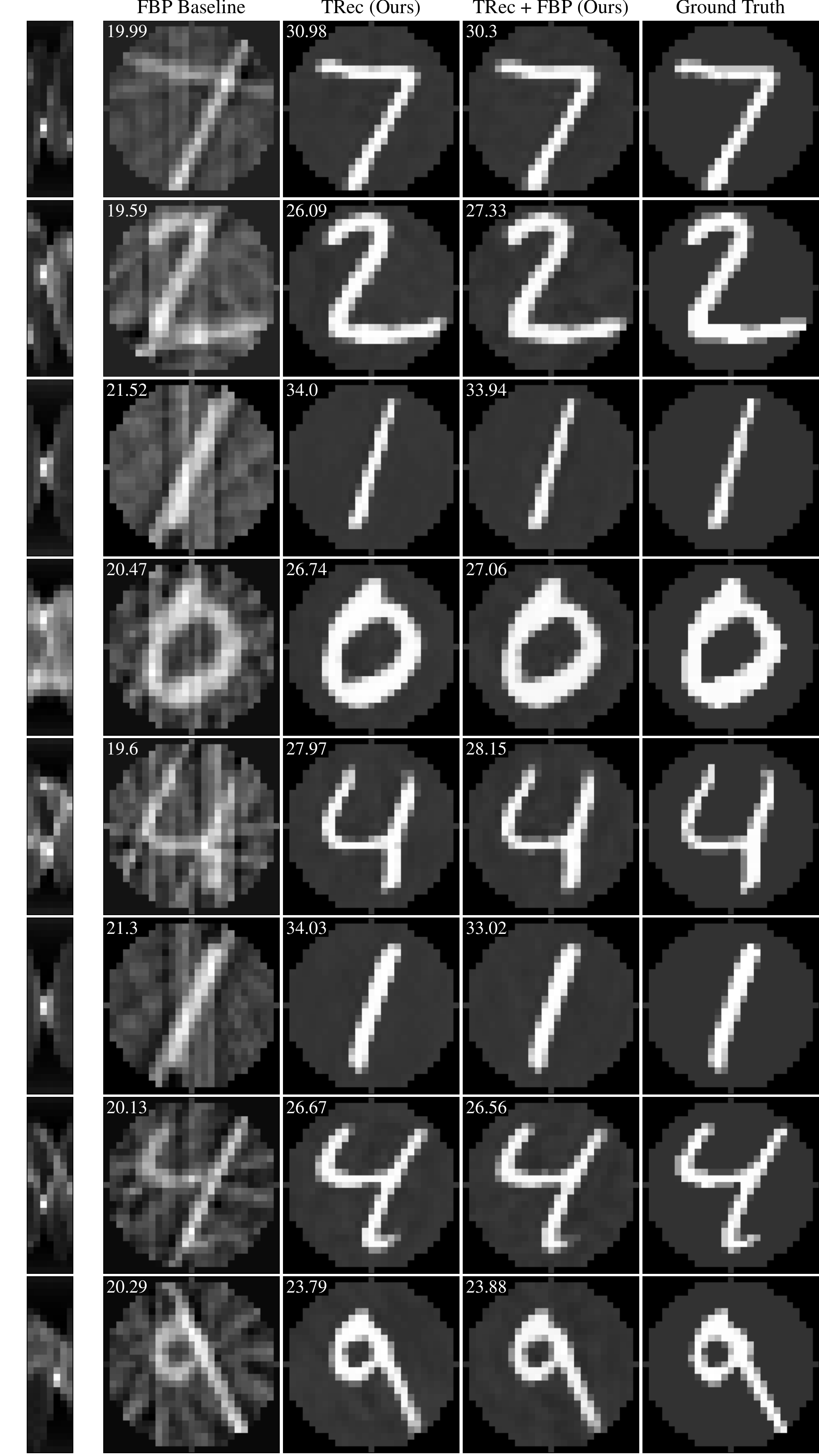}
    \put(0,8){\rotatebox{90}{\scriptsize{Input Sinograms ($7$ Projections, MNIST)}}}
    \end{overpic}
    
    \vspace{3pt}
    
    \begin{overpic}[width=.97\linewidth, trim=0 1083 0 21.5, clip,right]
    {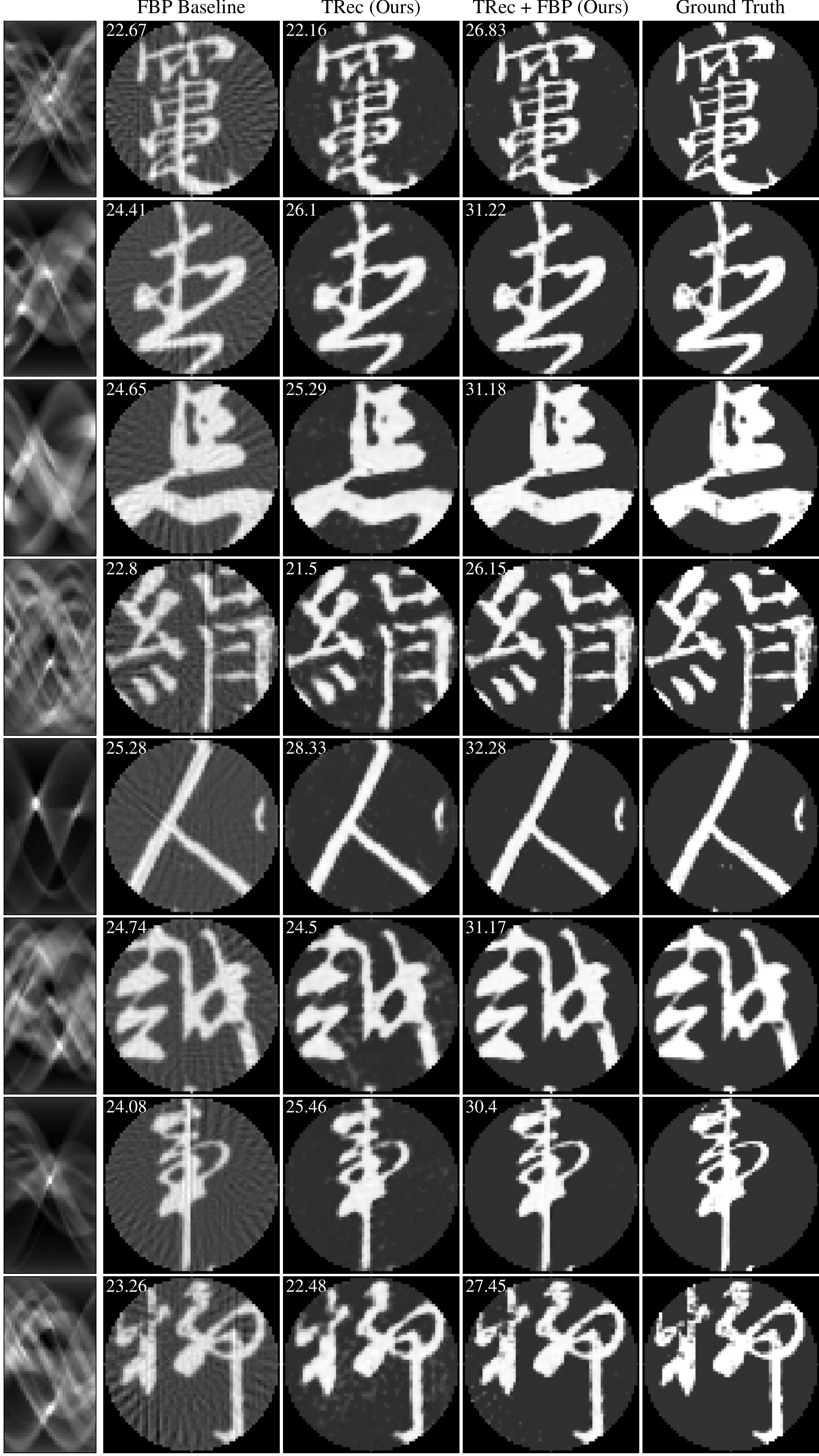}
    \put(0,8){\rotatebox{90}{\scriptsize{Input Sinograms ($33$ Projections, Kanji)}}}
    \end{overpic}
    
    \vspace{3pt}
    
    \begin{overpic}[width=.97\linewidth, trim=0 1083 0 21.5, clip,right]
    {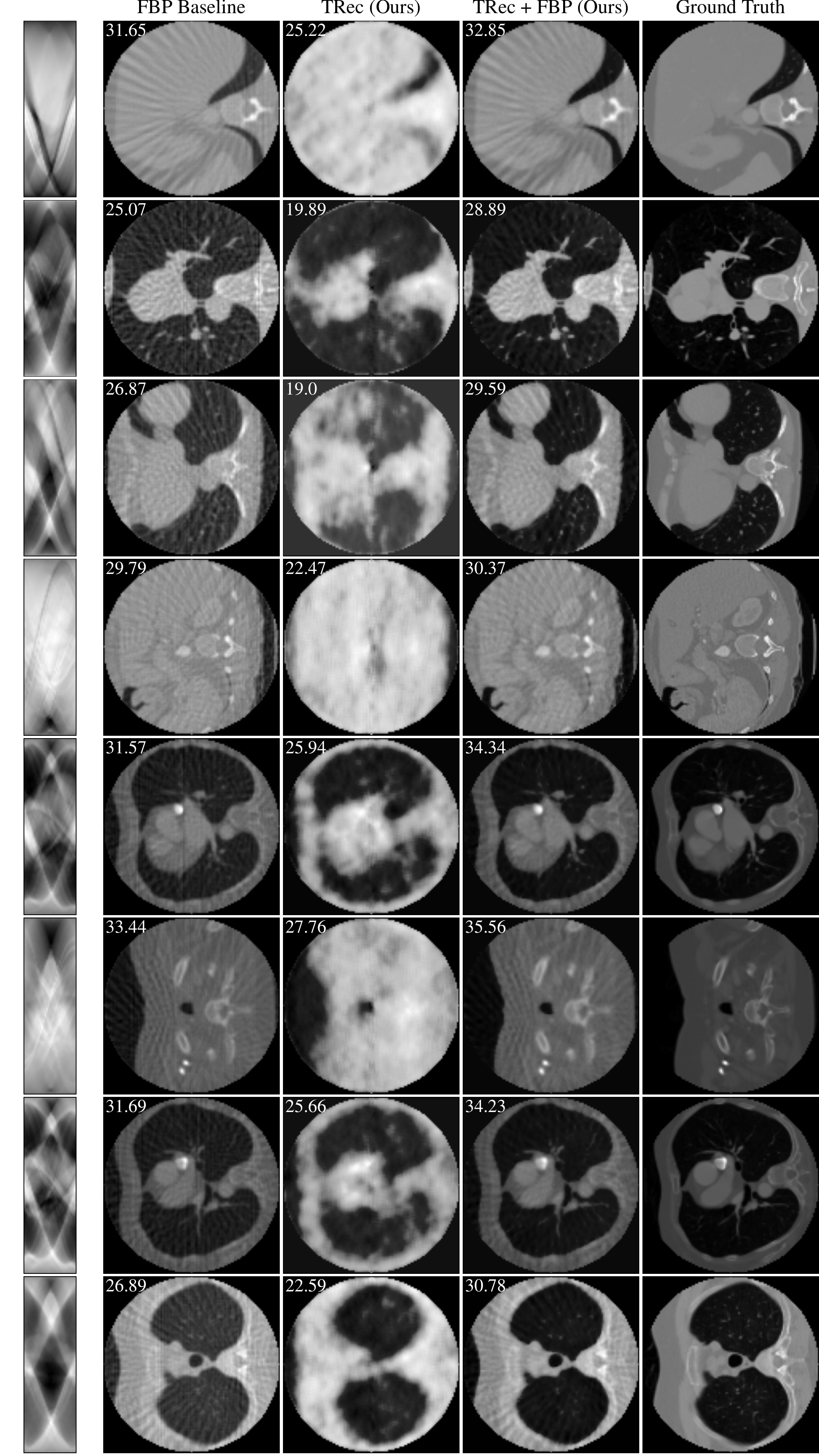}
    \put(0,6){\rotatebox{90}{\scriptsize{Input Sinograms ($33$ Projections, LoDoPaB)}}}
    \end{overpic}
    \caption{\textbf{Tomographic reconstruction results.}
    We show three qualitative results for each used datasets. 
    From left to right, we show the input sinogram, reconstruction results obtained with filtered backprojection (FBP)~\cite{kak2002principles,ramesh1989algorithm}, our results obtained with the ``FIT: TRec'' setup, our results obtained with the ``FIT: TRec + FBP'' setup, and the corresponding ground truth images. 
    In the top left corner of each reconstruction we show the peak signal-to-noise ratio (PSNR) with respect to the ground truth image.}
    \label{fig:trec_quali}
\end{figure}
}
\begin{document}

\title{Fourier Image Transformer}
\shorttitle{Fourier Image Transformer}

\author[1,\Letter]{Tim-Oliver Buchholz}
\author[2,\Letter]{Florian Jug}

\affil[1]{FMI for Biomedical Research, Basel, Switzerland}
\affil[2]{Fondazione Human Technopole, Milano, Italy}

\maketitle

\begin{abstract}
Transformer architectures show spectacular performance on NLP tasks and have recently also been used for tasks such as image completion or image classification.
Here we propose to use a sequential image representation, where each prefix of the complete sequence describes the whole image at reduced resolution.
Using such Fourier Domain Encodings (FDEs), an auto-regressive image completion task is equivalent to predicting a higher resolution output given a low-resolution input.
Additionally, we show that an encoder-decoder setup can be used to query arbitrary Fourier coefficients given a set of Fourier domain observations.
We demonstrate the practicality of this approach in the context of computed tomography (CT) image reconstruction.
In summary, we show that Fourier Image Transformer (FIT) can be used to solve relevant image analysis tasks in Fourier space, a domain inherently inaccessible to convolutional architectures.
\end{abstract}

\section{Introduction}
\label{sec:introduction}
Transformer architectures are currently setting new standards on virtually all natural language processing (NLP) tasks~\cite{devlin2018bert,radford2018improving}.
But also in other domains, transformers can find application~\cite{10.1093/bioinformatics/btz682}.
Vision Transformers (ViT)~\cite{dosovitskiy2020image}, for example, show SOTA results on image classification tasks.
In the future we expect an increasing number of problems to be solved with transformers and here we also contribute such applications.

The key novelty of transformers is their self-attention mechanism~\cite{vaswani2017attention}, allowing them to learn and utilize long ranging dependencies in data. 
Recently, this mechanism is applied to longer and longer input sequences of words or other elements, \eg pixels~\cite{chen2020generative}.

\figTeaser

Initially, we wondered if pixel sequences are the only valid image representation to train \textit{auto-regressive transformer} models in the spirit of~\cite{chen2020generative}.
In particular, we want to use a representation where each prefix of such a descriptive sequence encodes the full image at lower resolution.
Hence, we introduce \textit{Fourier Domain Encodings}~(FDEs), which do have this desired property and, as we show, can successfully be used to train auto-regressive Fourier Image Transformers (``FIT: SRes'') for super-resolution tasks (see Figures~\ref{fig:teaser} and~\ref{fig:sres_overview}). 

Additionally, we show how an \textit{encoder-decoder} based Fourier Image Transformer (``FIT: TRec'') can be trained on a set of Fourier measurements and then used to query arbitrary Fourier coefficients, which we use to improve sparse-view computed tomography (CT) image restoration\footnote{Sparse-view CT image restoration typically suffers from missing high frequency Fourier coefficients, leading to unwanted image artefacts.}. 
We demonstrate this by providing a given set of projection Fourier coefficients to our encoder-decoder setup and use it to predict Fourier coefficients at arbitrary query points.
This allows us to predict a dense, grid-sampled discrete Fourier spectrum of a high quality CT reconstruction (see Figures~\ref{fig:teaser} and~\ref{fig:trec_overview}).

In Section~\ref{sec:related} we review some of the related transformer and tomography literature.
In Section~\ref{sec:methods} we introduce our novel Fourier Domain Encoding (FDE) and training strategies for auto-regressive and encoder-decoder transformer models. In Section~\ref{sec:experiments} we present our experiments and results on multiple datasets.

\section{Related Work}
\label{sec:related}
Transformer architectures are revolutionizing neural language processing (NLP), replacing recurrent neural networks (RNNs) and long short-term memory (LSTM) architectures in virtually all NLP tasks~\cite{devlin2018bert,radford2018improving}.
The success of transformers in NLP has naturally raised the question if computer vision tasks might as well benefit from transformer-like attention.
Recent work on image classification~\cite{SASA2019,dosovitskiy2020image} and pixel-by-pixel image generation/completion~\cite{parmar2018image,chen2020generative,katharopoulos_et_al_2020} were among the first to successfully demonstrate the applicability of transformers in the image-domain.
In~\cite{parmar2018image}, for example, the first $n$ pixels of the flattened input image are used to condition a generative transformer setup that then predicts the remaining image in an auto-regressive manner.

Concurrently with our own work, Lee-Thorp~\etal~\cite{lee2021fnet} have proposed an interesting idea to combine Fourier-space image information with Transformer-based attention. While their approach is inspiring and leads to much improved results on a number of tasks on data other than images, our goals of super-resolution and tomographic reconstruction are of different nature and bear no overlap with their investigations.

\subsection{Attention is all you need}
Vaswani~\etal~\cite{vaswani2017attention} were the first to introduce transformers.
More specifically, they introduced an encoder-decoder structure, where the encoder maps an input encoding
$\mathbf{x} \in \mathbb{R}^{N \times F}$ 
into a continuous latent space 
$\mathbf{z} \in \mathbb{R}^{N \times F}$, 
with $N$ corresponding to the number of input tokens and $F$ representing the feature dimensionality per token.
This latent space embedding $\mathbf{z}$ is then given to the decoder, which generates an $M$ long output sequence
$\mathbf{y} \in \mathbb{R}^{M \times F}$ iteratively, element by element.
This auto-regressive decoding scheme means that the decoder generated the $i$-th output token while not only observing $\mathbf{z}$, but also all $i-1$ output tokens generated previously.

More formally, a transformer is a function 
$T: \mathbb{R}^{N \times F} \xrightarrow{} \mathbb{R}^{N \times F}$, 
represented by $L$ transformer layers 
\begin{equation}
    T_l(\mathbf{x}) = f_l(A_l(\mathbf{x}) + \mathbf{x}),
\end{equation}
with $A_l$ denoting a self-attention module and $f_l$ being a simple feed forward network.

In the self-attention module, the input $\mathbf{x}$ is mapped to queries $Q = \mathbf{x}W_Q$, keys $K = \mathbf{x}W_K$ and values $V = \mathbf{x}W_V$ by matrix multiplication with learned matrices 
$W_Q \in \mathbb{R}^{F \times D}$, $W_K \in \mathbb{R}^{F \times D}$ 
and 
$W_V \in \mathbb{R}^{F \times F}$. 
The self-attention output is then computed by
\begin{equation}
\label{eq:softmax_attention}
    \begin{split}
        A_l(\mathbf{x}) &= \text{softmax}\left(\frac{QK^T}{\sqrt{D}}\right)V,
    \end{split}
\end{equation}
with the softmax-function being applied per row.
Intuitively, the softmax-normalized similarity between computed keys and queries is used to obtain the weighted sum over the values.

Typically, instead of a single self-attention module, multi-head attention is being used. 
If that is the case, a transformer layer $T_l$ learns multiple $W_Q$, $W_K$ and $W_V$, allowing the layer to simultaneously perform multiple attention-based computations~\cite{vaswani2017attention}.

Since transformers do not explicitly encode the relative position between input tokens, positional encodings are required whenever specific input topologies need to be made accessible to the transformer.
In~\cite{vaswani2017attention}, a useful 1D positional encoding scheme was proposed. 
Later, Wang~\etal~\cite{wang2020translating} generalized this scheme to 2D topologies.
In our own work, we have adopted this encoding scheme, but use it not only to encode integer pixel-grid locations, but arbitrary coordinates.

While the advantage of transcending beyond a CNN's localized receptive field by introducing global attention proves beneficial for many learning tasks, the big downside is the computational cost that comes with it.
Due to the required matrix multiplication $QK^T$, the self-attention on an input sequence of length $N$ requires $\mathcal{O}(N^2)$ memory and time.
Since image-based applications have to deal with very long input sequences, the use of efficient transformer implementations~\cite{kitaev2020reformer,wang2020linformer,bello2021lambdanetworks} is essential.
Hence, we use fast-transformers, an efficient approximation of softmax self-attention, introduced by Katharopoulos~\etal~\cite{katharopoulos_et_al_2020}.

\subsection{Tomographic Image Reconstruction}
\label{subsec:tomorec}
In computed tomography (CT), the radon transform~\cite{Radon17,kak2002principles} of a 2D section is acquired by rotating a 1D detector array around it, acquiring a series of density measurements at projection angles $\alpha_i$.
Such a series of acquisitions is often depicted as a 2D image, a so called sinogram, in which each pixel column corresponds to one such 1D measurement.
A widely used tomographic reconstruction method is filtered backprojection~(FBP)~\cite{kak2002principles, ramesh1989algorithm} (see Figure~\ref{fig:trec_overview}).
FBP is based on the Fourier slice theorem, which states that the Fourier coefficients of each 1D projection at a given angle $\alpha_i$ coincide with the 2D Fourier coefficients that lie on the line that crosses through the DC component of the Fourier space representation at angle $\alpha_i$~\cite{bracewell1956strip}.
Hence, we can compute the 1D Fourier transformation of the 1D projection measurements and arrange them according to their projection angle $\alpha_i$ in 2D Fourier space, take the inverse Fourier transformation and thereby obtain the reconstructed 2D image.

In practice, it is desirable to limit the number of projections in order to reduce overall acquisition times and total sample exposure.
However, having fewer measurements leads wedges of missing Fourier measurements, which then lead to reconstruction artefacts. 
More specifically, such missing wedges lead to radial striping artefacts in reconstructed images, best observed around image locations that are subject to large intensity gradients (see \eg Figure~\ref{fig:trec_overview}).

\figSResOverview

A plethora of methods to reduce these artefacts was proposed~\cite{chen2017low,jin2017deep,adler2018learned,hauptmann2018model}, but none of them is operating in Fourier space.
Below, we will introduce Fourier Domain Encodings (FDEs) which is based on the Fourier transformation of an image. 
Then we train a transformer to fill in all unobserved Fourier coefficients such that missing-wedge artefacts will be reduced or, ideally, avoided.

\section{Methods}
\label{sec:methods}

\figTRecOverview

\subsection{Fourier Domain Encodings (FDEs)}
\label{sec:fde}
To compute the Fourier Domain Encoding (FDE) for an image $\bm{x}$, we first take the discrete Fourier transform (DFT), $\bm{X}=\mathcal{F}(\bm{x})$, giving us the complex valued Fourier spectrum $\bm{X}$. 
The DC component of $\bm{X}$ is in its center-most location.
Concentric rings of Fourier coefficients around the DC component are called Fourier rings.
More central Fourier rings contain lower frequencies, coefficients further away from the DC component higher ones. 
Since we started with a real-valued image $\bm{x}$, we know that the Fourier spectrum is radially symmetric, which allows us to drop half of $\bm{X}$, resulting in $\bm{X}_h$.
A lower resolution real image $\bm{x_{r}}$ of the original image $\bm{x}$ can be obtained by masking all Fourier coefficients above a given radius $r$ and computing the inverse Fourier transform of the result. 
Effectively the real image resolution depends entirely on the value of $r$, see Figure~\ref{fig:sres_overview} for an example.

For the Fourier Domain Encoding (FDE) we convert the complex Fourier coefficient rings into a 1D sequence starting form the DC component followed by an unrolling of the (half) Fourier rings from $\bm{X}_h$
\begin{equation}
    \mathbf{S} = \text{unroll}(\bm{X}_h) = \left[c_1, c_2, \dots, c_N\right]^T,
\end{equation}
where $c_i$ -- the words of our input sequence -- correspond to the complex Fourier coefficients.
The sequence $\mathbf{S}$ has the interesting property that any sub-sequence $\mathbf{S}_j$ starting at $c_1$ and ending at $c_j, j \in [2, N]$ encodes the whole real image at lower resolutions.

In order to proceed, we convert the complex Fourier coefficients $c_i$ into normalized amplitudes 
\begin{equation}
    a_i = \frac{2(|c_i| - a_{max})}{a_{max} - a_{min}} - 1
\end{equation}
and phases
\begin{equation}
    \phi_i = \frac{\angle(c_i)}{\pi},
\end{equation}
where $a_{min}$ and $a_{max}$ are minimum and maximum amplitudes computed over all training images and the function $\angle$ returns the phase of a given Fourier coefficient.
Hence, the complex sequence $\mathbf{S}$ can now be described by the normalized real-valued matrix
\begin{gather}
    \mathbf{C} = \begin{bmatrix} 
                    a_1 & \cdots & a_N \\
                    \phi_1 & \cdots &  \phi_N
                 \end{bmatrix}^T
\end{gather} 
with $\mathbf{C} \in \mathbb{R}^{N \times 2}$.

Our final goal is to transform each word $(a_i, \phi_i)$ into an $F$-dimensional vector.
To this end we feed $\mathbf{C}$ through a single trainable linear layer that increases the feature dimensionality from $2$ to $\frac{F}{2}$, to which we concatenate an $\frac{F}{2}$-dimensional 2D positional encoding\footnote{
We slightly adapted the 2D positional encoding of~\cite{wang2020translating} to accept arbitrary (non integer) coordinates.} 
that represents the original polar coordinates of the Fourier coefficient in the original 2D Fourier spectrum $\bm{X_h}$.
The final FDE image sequence is therefore $\bm{E}\in \mathbb{R}^{N \times F}$.
The sub-sequence property of $\bm{S}_j$ holds for $\bm{E}_j$ as well.

Predicted output words $\mathbf{Z}=[z_1,\dots,z_k]$, with $z_i \in \mathbb{R}^F$, are fed through two linear layers that back-transform the $F$-dimensional encoding of $z_i$ into predicted amplitudes and phases $\hat{c}_i = (\hat{a}_i, \hat{\phi}_i)$, respectively.
The output of the phase-predicting layer is additionally passed through the $tanh$-activation function to ensure that all phases are in $[-1, 1]$.

\figSResQuant
\figSResCond

\subsection{Fourier Coefficient Loss}
\label{sec:fc_loss}
We train our Fourier Image Transformers (FITs) with a loss function consisting of two terms, 
$(i)$~the amplitude loss 
\begin{equation}
    \label{eq:amp_loss}
    \mathcal{L}_{amp}(\hat{a}_i, a_i) = 1 + (\hat{a}_i - a_i)^2,
\end{equation}
computed between the predicted amplitudes $\hat{a}_i$ and the target amplitudes $a_i$, and
$(ii)$~the phase loss
\begin{equation}
    \label{eq:phase_loss}
    \mathcal{L}_{\angle}(\hat{\phi}_i, \phi_i) = 2 - cos(\hat{\phi}_i - \phi_i),
\end{equation}
with $\hat{\phi}_i$ the predicted phase and $\phi_i$ the corresponding target phase.

The final Fourier coefficient loss $\mathcal{L_{FC}}$ is the multiplicative  combination of both individual losses, given by
\begin{equation}
\label{eq:fc_loss}
    \mathcal{L_{FC}}(\hat{\mathbf{C}}, \mathbf{C}) = \frac{1}{N}\sum_{i=0}^{N} \mathcal{L}_{amp}(\hat{a}_i, a_i) \cdot \mathcal{L}_{\angle}(\hat{\phi}_i, \phi_i).
\end{equation}

\subsection{FIT for Super-Resolution}
\label{sec:method_sres}
The sub-sequence property of the Fourier domain encoding (FDE) $\mathbf{E}$, where each $\mathbf{E}_j$ encodes a lower resolution image $\bm{x}_{j}$ enables us to train an auto-regressive Fourier Image Transformer for super-resolution (``FIT: SRES'').

This FIT for super-resolution (``FIT: SRes'') gets an FDE sequence $\bm{E}_{N-1}=[e_1, \dots, e_{N-1}]$ as input and is trained \wrt the correct target sequence $\bm{Z} = [z_2, \dots, z_{N}]$, using the previously introduced Fourier coefficient loss $\mathcal{L_{FC}}$, computed on the back-transformed predicted amplitude and phase values $\bm{\hat{C}}=[\hat{c}_2,\dots,\hat{c}_N]$, where $\hat{c}_i=(\hat{a}_i, \hat{\phi}_i)$, as explained in Section~\ref{sec:fde}.

Once the transformer is trained, a sub-sequence $\bm{E}_j = [e_1, \dots, e_{j}], j \in [2, N]$ is used to condition the transformer, which is then used in iterations to auto-regressively predict the missing part of the complete sequence $\bm{E} = [e_1, \dots, e_N]$, \ie filling in predicted high-frequency information not contained in $\bm{E}_j$ (see Figure~\ref{fig:sres_overview}).

Note that the proposed super-resolution setup operates exclusively on Fourier domain encoded data.
All final prediction images $\bm{\hat{x}}$ are generated by computing the inverse Fourier transform on predictions $\bm{\hat{C}}$, which are rearranged (rolled) into $\bm{\hat{X}_h}$ and completed to a full predicted Fourier spectrum $\bm{\hat{X}}$, \ie $\bm{\hat{x}} = \mathcal{F}^{-1}(\text{roll}(\bm{\hat{C}}))$.

\subsection{FIT for Tomography}
\label{sec:tomorec}
Our Fourier Image Transformer setup for tomographic reconstruction (``FIT: TRec'') is based on an encoder-decoder transformer architecture as shown in Figure~\ref{fig:trec_overview}.

As input to the encoder we use the Fourier Domain Encoding (FDE) of a raw sinogram $\bm{s}$.
As described above, $\bm{s}$ consists of $P$ pixel columns $[s_1,\dots,s_P]$ of 1D projections of $\bm{x}$ at angles $[\alpha_1,\dots,\alpha_P]$.
The Fourier slice theorem states, see also Section~\ref{subsec:tomorec}, that the discrete 1D Fourier coefficients $\bm{C}_i=\mathcal{F}(s_i)$ coincide with the values of the 1D slice at angle $\alpha_i$ through the 2D Fourier spectrum $\mathcal{F}(\bm{x})$.
To assemble the full FDE of a sinogram we need to combine all $\bm{C}_i$ with the adequate positional encoding (using polar coordinates) of all Fourier coefficients, as dictated by the Fourier slice theorem and sketched in Figure~\ref{fig:trec_overview}.

Hence, the encoder creates a latent space representation $\bm{Z}$ that encodes the full input sinogram $\bm{s}$.
This latent space encoding is then given as input to the decoder.
The decoder is then used to predict all Fourier coefficients $\bm{\hat{C}}$, such that the predicted reconstruction $\bm{\hat{x}}$ of $\bm{x}$ can be computed by $\bm{\hat{x}} = \mathcal{F}^{-1}(\text{roll}(\bm{\hat{C}}))$, where roll arranges the 1D sequence back into a discrete 2D Fourier spectrum.
We call this setup ``FIT: TRec''.

Additionally, we propose a variation of this procedure, called ``FIT: TRec + FBP'', where the decoder not only receives the latent space encoding $\bm{Z}$, but also FDEs of the Fourier coefficients $\bm{C}_{\text{FBP}}=\mathcal{F}(\text{FBP}(\bm{s}))$, where FBP denotes the function computing the filtered backprojection of a sinogram (see Figure~\ref{fig:trec_overview}).

Note that the implementation of ``FIT: TRec'' coincides with ``FIT: TRec + FBP'', with FBP being replaced by a function ZERO which returns $0$ for all inputs.

We train ``FIT: TRec'' and ``FIT: TRec + FBP'' using the $\mathcal{L_{FC}}$-loss of Eq.\ref{eq:fc_loss}.
Additionally, we introduced a residual convolution block consisting of two convolutional layers ($3 \times 3$ followed by $1 \times 1$) with $d_{\text{conv}}=8$ intermediate feature channels.
This conv-block (conv) receives the inverse Fourier transform of the predicted Fourier coefficients $\bm{\hat{x}}=\mathcal{F}^{-1}(\text{roll}(\bm{\hat{C}}))$ as input and is trained using the MSE-loss between the predicted real-space image $\text{conv}(\bm{\hat{x}})$ and the known ground truth image $\bm{x}$.
Hence, the full loss is the sum over $\mathcal{L_{FC}}$ and the MSE-loss.

In order to speed up training, we start by feeding only a low-resolution subset of all Fourier coefficients $\bm{C}_i=\mathcal{F}(s_i)$ (and $\bm{C}_{\text{FBP}}$), and successively increase this subset over training until the full sets are used.
This forces the FIT to first learn good low resolution features and later learn to add suitable high resolution predictions.

\section{Experiments and Results}
\label{sec:experiments}
\label{sec:results}

Here we describe the experiments we conduct with the previously described super-resolution (``FIT: SRes'') and tomographic reconstruction (``FIT: TRec'' and ``FIT: TRec + FBP'') setups.
Note, all code used to create results reported in this manuscript is available on GitHub\footnote{\url{https://github.com/juglab/FourierImageTransformer}}.

\subsection{Data and Metrics for Super-Resolution}
\label{sec:data_sres}
\compactsubsub{MNIST~\cite{lecun-mnisthandwrittendigit-2010}} 
Original images are cropped to $27 \times 27$ pixels with the default PyTorch train-test split (in $60'000$ and $10'000$ images, respectively). 
The train images are further split into $55'000$ samples for training and $5'000$ validation images.

\compactsubsub{CelebA $\mathbf{128 \times 128}$~\cite{liu2015faceattributes}}
Original images are converted to gray scale and downscaled to $63 \times 63$ pixels. 
The images are randomly split into $20'000$, $5'000$ and $5'000$ training, validation and test samples, respectively.

We evaluate all results using 
$(i)$~Fourier Ring Correlation (FRC)~\cite{van1982arthropod} in Fourier space and 
$(ii)$~the peak signal-to-noise ratio (PSNR) in image space.

\subsection{Data and Metrics for Computed Tomography}
\label{sec:data_trec}
\compactsubsub{Data Preparation for Tomographic Reconstruction}
As described in Section~\ref{subsec:tomorec}, tomographic image reconstruction in 2D operates on a number of 1D projections of a given true object $\bm{x}$.
Our tomographic simulation process is based on the work by Leuschner~\etal~\cite{leuschner2019lodopab}.
Furthermore, we chose the detector length to be equal to the width of our chosen object (\ie ground truth image) we apply our synthetic tomography pipeline on.
To avoid spurious contributions to individual projections, we need to set all pixel intensities outside the largest image-centered circle to $0$ (hence, we see only circular images in Figures~\ref{fig:trec_overview}, and \ref{fig:trec_quali}).

\compactsubsub{MNIST~\cite{lecun-mnisthandwrittendigit-2010}} Data is split and preprocessed as described in Section~\ref{sec:data_sres}.
Additionally, for visualization purposes, we min-clip all pixel intensities within the before-mentioned largest image-centered circle to $50$.
Finally, we use this data to compute $P=7$ equally spaced projections which we assemble in sinograms $\bm{s}^j_{\text{MNIST}}=[s^{j,1}_{\text{MNIST}},\dots,s^{j,P}_{\text{MNIST}}]$.

\compactsubsub{Kanji~\cite{clanuwat2018deep}} 
Data is randomly split into $50'000$ train, $5'000$ validation and $5'000$ test samples and all images are cropped to $63 \times 63$ pixels, which are otherwise processed as described for MNIST.
Finally, we use this data to compute $P=33$ equally spaced projections which we assemble in sinograms $\bm{s}^j_{\text{Kanji}}=[s^{j,1}_{\text{Kanji}},\dots,s^{j,P}_{\text{Kanji}}]$.

\compactsubsub{LoDoPaB~\cite{leuschner2019lodopab}} 
The original train- and validation-data is first reduced to $4'000$ and $400$ randomly chosen images respectively and for testing all $3'553$ images are used.
All selected images are downscaled to $111 \times 111$ pixels and we compute $P=33$ equally spaced projections like for the Kanji data.

All tomographic reconstruction experiments with ``FIT: TRec'' and ``FIT: TRec + FBP'' and the FBP baseline are evaluated using peak signal-to-noise ratio (PSNR) \wrt available ground truth.

\subsection{Training Setup for Super-Resolution}
We use a $F=256$ dimensional FDE, with the positional encoding being based on polar coordinates, \ie Fourier coefficients of same frequency have the same radius.
The FDE is passed to a \textit{causal-linear} transformer~\cite{katharopoulos_et_al_2020} with $8$ layers, $8$ self-attention heads, a query and value dimensionality of $32$, dropout of $0.1$, attention dropout of $0.1$, and a dimensionality of the feed-forward network of $1024$.

This setup is trained auto-regressively, \ie with a triangular attention mask.
We use the rectified Adam optimizer (RAdam)~\cite{liu2019variance} with an initial learning rate of $0.0001$ and weight decay of $0.01$ for $100$ epochs.
The batch size is $32$.
We half the learning rate on plateauing validation loss.

\subsection{Results for Super-Resolution Tasks}
Quantitative results for all conducted super-resolution experiments on MNIST data are shown in Figure~\ref{fig:sres_quant}, where we show 
$(i)$~$3 \times$ binned low-res MNIST input images corresponding to $\bm{E}_{pre} = [c_1, \dots, c_{39}]$, also sketched in Figure~\ref{fig:sres_overview},
$(ii)$~corresponding ground truth MNIST images,
$(iii)$~the predictions of the ``FIT: SRes'' network trained on the MNIST data as described in Section~\ref{sec:method_sres},
$(iv)$~two box-plots showing the distribution of PSNR values computed between the ground truth images and the downscaled inputs and predicted outputs, respectively, and
$(v)$~Fourier ring correlation plots showing the correlation between our predicted Fourier coefficients and the corresponding ground truth.

In Figure~\ref{fig:sres_cond}, we show $3$ sequences of super-resolution results obtained with a FIT trained on the CelebA data.
For each image, we conditioned the trained transformer on $2$, $4$, $8$, and $16$ Fourier rings, respectively. 
This corresponds to low-resolution images subject to $16\times$, $8\times$, $4\times$, and $2\times$ binning in Fourier space, respectively.

\subsection{Training Setup for Computed Tomography}
Like before, we consistently use $F=256$ dimensional FDEs, and employ the \textit{linear} encoder and decoder method by Katharopoulos~\etal~\cite{katharopoulos_et_al_2020}.
More specifically, we use $4$ transformer layers, $8$ self-attention heads per layer, a query and value dimensionality of $32$, dropout of $0.1$, attention dropout of $0.1$, and a dimensionality of the feed-forward network of $1024$.
The residual conv-block has $d_{\text{conv}}=8$ intermediate feature channels.

All networks are optimized using RAdam~\cite{liu2019variance}, with an initial learning rate of $0.0001$ and weight decay of $0.01$ for $300$ (MNIST), $120$ (Kanji), and $350$ (LoDoPaB) epochs. 
The batch size is $32$.
We half the learning rate on plateauing validation loss.

\begin{table}[t]
    \centering
    \begin{tabular}{l|l|c}
        Dataset & Method & PSNR \\
        \hline
        \hline
        \multirow{5}{*}{MNIST~\cite{lecun-mnisthandwrittendigit-2010}} & Baseline: FBP & $17.87$\\
                               & FIT: TRec + FBP (Ours) & $27.85$ \\ 
                               & FIT: TRec (Ours)  & $\mathbf{27.90}$ \\ \cline{2-3} 
                               & Ablation: Only FBP & $26.89$ \\
                               & Ablation: Only Conv-Block & $22.53$ \\ 
        \hline \hline
        \multirow{5}{*}{Kanji~\cite{clanuwat2018deep}} & Baseline: FBP & $22.06$\\
                               & FIT: TRec + FBP (Ours) & $\mathbf{30.72}$ \\ 
                               & FIT: TRec (Ours)  & $25.99$ \\ \cline{2-3} 
                               & Ablation: Only FBP & $30.49$ \\
                               & Ablation: Only Conv-Block & $26.92$ \\ 
        \hline \hline
        \multirow{5}{*}{\shortstack[l]{LoDoPaB~\cite{leuschner2019lodopab}\\ \small{(downscaled)}}} & Baseline: FBP & $26.89$\\
                               & FIT: TRec + FBP (Ours) & $\mathbf{30.98}$ \\ 
                               & FIT: TRec (Ours)  & $21.90$ \\ \cline{2-3} 
                               & Ablation: Only FBP & $30.74$ \\
                               & Ablation: Only Conv-Block & $30.70$ \\ 
        \hline
    \end{tabular}
    \vspace{2mm}
    \caption{\textbf{Quantitative tomographic reconstruction results.}
    We report the average peak signal-to-noise ratio (PSNR) with respect to ground truth, for each of the three used datasets.
    For each dataset, we compare the results of our ``FIT: TRec + FBP'' and ``FIT: TRec'' setups to results obtained with the filtered backprojection (FBP)~\cite{kak2002principles,ramesh1989algorithm} baseline, and the two ablation studies described in Section~\ref{sec:ablation}.}
    \label{tab:trec_res}
\end{table}

\figTRecQuali

\compactsubsub{Ablation Studies}
\label{sec:ablation}
We propose two ablation setups for all tomographic reconstruction experiments.

First, we ask what influence the encoded latent space information $\bm{Z}$, \ie the output of the encoded sinogram, has on the quality of the overall reconstruction $\bf{\hat{x}}$.
To that end, we perform ablation experiments for all $3$ datasets, for which we do not feed $\bm{Z}$ to the decoder. Technically this is implemented by replacing the decoder by an encoder network (since only one input remains to be fed).
We label these experiments ``Only FBP''.

The second ablation study asks, to what degree the conv-block contributes to the overall reconstruction performance, \ie we want to verify that the convolution block alone is not sufficient to solve the task at hand.
Hence, we train the conv-block on pairs of images $(\text{FBP}(\bm{s}),\bm{x})$, \ie the filtered backprojection of sinograms $\bm{s}$ and their corresponding ground truth images $\bm{x}$.
We label these experiments ``Only Conv-Block''.

For all ablation experiments all hyper-parameters not explicitly mentioned above are kept unchanged.

\subsection{Results for Tomographic Reconstruction Tasks} 
Qualitative tomographic reconstruction results for all three datasets we used are shown in Figure~\ref{fig:trec_quali}.
For each dataset, we show three input sinograms, the reconstruction baseline obtained via filtered backprojection~(FBP), our results obtained via ``FIT: TRec''and ``FIT: TRec + FBP'', and the corresponding ground truth images.

In Table~\ref{tab:trec_res}, PSNR numbers for all three datasets using the FBP baseline, our ``FIT: TRec + FBP'' and ``FIT: TRec'' training setups, and both ablation studies are given.

\section{Discussion}
\label{sec:discussion}
We proposed the idea of Fourier Domain Encodings (FDEs), a novel sequential image encoding, for which each prefix represents the whole image at reduced resolution, and demonstrated the utility of FDEs for solving two common image processing tasks with transformer networks, \ie super-resolution and tomographic image reconstruction. 

For the super-resolution task we showed that Fourier Image Transformer can be trained such that, when conditioned on an FDE corresponding to a low-resolution input image, can auto-regressively predict an extended FDE sequence that can be back-transformed into a higher resolution output.
It is obvious, the information required to generate a higher resolution image must be stored in the trained network, and we have shown in Figure~\ref{fig:sres_cond}, how this learned prior completes very low to moderate resolution inputs in sensible ways.
It is curious to see that eyes are the first high-resolution structures filled in by the trained FIT.
We believe that this is a direct consequence of all training images being registered such that the eyes are consistently at the same location. 

For the tomographic reconstruction task, we employ an encoder-decoder transformer that encodes a given FDE sequence corresponding to a given sinogram and can then be used to predict Fourier coefficients at arbitrary query locations.
We used the decoder to predict all Fourier coefficients of a fully reconstructed image, which we could then visualize via inverse Fourier transformation (iFFT).
We noticed that introducing a shallow residual convolution block after the iFFT reduces unwanted high frequency fluctuations in predicted results.
While we see that this procedure leads to very convincing results on MNIST, for more complex datasets, results quickly deteriorate. 
Hence, we proposed to additionally feed Fourier coefficients obtained by filtered backprojection (FBP) into the decoder.
This leads to much improved results that outperform the FBP baseline, showing that the FIT does contribute to solving the reconstruction task.

Our results show that transformers, currently the dominant approach for virtually all NLP tasks, can successfully be applied to complex and relevant tasks in computer vision.
While this is encouraging, we see a plethora of possibilities for future improvements.
For example, the transformers we used are rather small. 
We believe that an up-scaled version of our training setups with more attention heads and more layers would already lead to much improved results\footnote{\url{https://tinyurl.com/needadviceforafriend}}.
Still, we also believe that there is plenty of room for methodological improvements that do not require more computational resources, making this line of research also accessible to many other research labs around the globe.

\section*{Acknowledgments}
The authors would like to thank Mangal Prakash and Manan Lalit for discussions and help to improve the text. We also want to thank the Scientific Computing Facility at MPI-CBG for giving us access to their HPC infrastructure.
Funding was provided from the Max-Planck Society under project code M.IF.A.MOZG8106, the core budget of the Max-Planck Institute of Molecular Cell Biology and Genetics (MPI-CBG), the Human Technopole, and the BMBF under codes 031L0102 (de.NBI) and 01IS18026C (ScaDS2), as well as by the Deutsche Forschungsgemeinschaft (DFG) under code JU3110/1-1 (FiSS) and TO563/8-1 (FiSS). 
Additionally, we thank Yannic Kilcher and his YouTube channel\footnote{\url{https://www.youtube.com/c/YannicKilcher}} which greatly facilitated our desire to think about applications of transformers on images.

\section{Bibliography}
\bibliography{refs.bib}
\end{document}